\documentclass[conference,a4paper]{IEEEtran}
\usepackage{cite}
\usepackage{amsmath,amssymb,amsfonts}
\usepackage{graphicx}
\usepackage{textcomp}
\usepackage{xcolor}
\usepackage{multirow,booktabs}
\usepackage{array,tabularx}
\usepackage[caption=false]{subfig}
\usepackage{placeins}
\usepackage{url}
\graphicspath{{../}}

\newcommand{\dyaw}{\Delta\psi}
\newcommand{\dyawt}{\Delta\psi_t}
\newcommand{\Ours}{\textsc{Ours}}
\newcommand{\best}[1]{\textbf{#1}}
\newcommand{\tabhead}[1]{\shortstack{#1}}
\newcolumntype{Y}{>{\raggedright\arraybackslash}X}
\renewcommand\IEEEkeywordsname{Keywords}

\newcommand{\NUDTauthor}[1]{%
\begin{minipage}[t]{0.305\textwidth}
\centering
{\footnotesize
#1\\[0.35ex]
\textit{College of Intelligence Science and Technology}\\
\textit{National University of Defense Technology}\\
Changsha, China\par}
\end{minipage}}

\begin{document}

\title{LP-NavOA: Integrated Local Navigation and Obstacle Avoidance
for Humanoid Robots under Limited Perception}

\author{%
\setlength{\tabcolsep}{0pt}%
\begin{tabular*}{0.98\textwidth}{@{\extracolsep{\fill}}ccc@{}}
\NUDTauthor{1\textsuperscript{st} Yukun Luo} &
\NUDTauthor{2\textsuperscript{nd} Jianjun Ma} &
\NUDTauthor{3\textsuperscript{rd} Yuyao Min} \\[4.6em]
\NUDTauthor{4\textsuperscript{th} Jinzhe Li} &
\NUDTauthor{5\textsuperscript{th} Kaihong Huang} &
\NUDTauthor{6\textsuperscript{th} Peng Li}
\end{tabular*}%
}

\maketitle
\vspace{-2.2em}

\begin{abstract}
Humanoid local navigation in cluttered environments must jointly resolve
obstacle avoidance, sparse-goal recovery, and stable whole-body
locomotion under short-range and partially observable sensing. Explicit
planner-control decompositions introduce latency and can mismatch agile
humanoid command-tracking limits, while purely reactive controllers may
lose the goal after obstacle occlusion. We present LP-NavOA, a
limited-perception navigation and obstacle-avoidance framework for
humanoid robots. A raycast-conditioned perception-action proximal policy
optimization (PPO) locomotion backbone is
first trained with a robot-centered circular heading-speed command and a
shared command-side safety filter. With this backbone frozen,
A-star and waypoint teachers generate rollouts for distilling a
recurrent local planner that overwrites only the heading command at
deployment, leaving the whole-body policy intact. At runtime, LP-NavOA uses
proprioception, short-range local range sensing, and a body-frame goal
direction, requiring no global map, waypoint stream, or external
planner. In MuJoCo open-wall and indoor layouts, the distilled planner
produces obstacle bypassing and post-avoidance goal recovery, raising
teacher-calibrated on-time arrival from 38--40\% to 85--97\% and reducing
brush/contact-heavy progress relative to a backbone-only controller.
Ablations show that dynamic route shaping, teacher-active data
collection, and the circular command interface are important for
navigation efficiency and for training the 3.0\,m/s backbone. A Unitree
G1 deployment analysis demonstrates hardware executability without continuous
joystick steering.
\end{abstract}


\section{Introduction}
\label{sec:intro}

Humanoid robots fit human-scale spaces, but local navigation is tightly
coupled with whole-body locomotion. A policy must decide when to walk,
turn, slow down, or return toward a hidden goal while preserving balance
and contact quality. Under partial observability~\cite{kaelbling1998pomdp},
short-range sensing may reveal an obstacle without resolving when to
continue bypassing or resume goal seeking.

Classical pipelines separate mapping, planning, and control with
explicit planners such as A-star and dynamic
windows~\cite{hart1968astar,fox1997dwa}. These interfaces are
interpretable but awkward for agile legged systems, where simplified
planning abstractions meet latency, body sway, and command-tracking
limits. Learned navigation systems have reduced some of these
costs~\cite{kahn2021badgr,shah2023vint,sridhar2024nomad}, and recent
work extends the same trend to humanoid visual
navigation~\cite{huang2026tnavrl}. In parallel, reinforcement learning
has produced robust legged locomotion under increasingly realistic
dynamics, from agile motor skills~\cite{hwangbo2019legged} to blind
locomotion over challenging terrain~\cite{lee2020legged} and perceptive
locomotion in the wild~\cite{miki2022wild}. Rapid motor adaptation
pushes this line further by handling unmodelled dynamics directly at
deployment~\cite{kumar2021rma}, a property that humanoid systems have
begun to inherit alongside real-world whole-body
control~\cite{radosavovic2024humanoid}, expressive whole-body
tracking~\cite{cheng2024exbody}, and large-scale imitation from human
motion~\cite{fu2024humanplus}. Scalable simulators such as Isaac Gym
have made massively parallel training practical~\cite{makoviychuk2021isaacgym},
with curricula that learn to walk in minutes~\cite{rudin2022leggedgym};
lightweight MuJoCo-based systems such as mjlab further expose
manager-based robot-learning workflows for GPU-accelerated policy
training~\cite{zakka2026mjlab}. Building on this style of simulation
infrastructure, recent studies pursue collision-free high-speed
locomotion~\cite{he2024agile}, extreme parkour~\cite{cheng2024extreme},
and rapid-locomotion benchmarks~\cite{margolis2024rapid}. Yet
goal-directed local navigation still requires recovering the sparse
goal after avoidance, not only reacting to the nearest obstacle.

We separate \emph{training instruments} from the \emph{inference
interface}. Planners, privileged scans, and waypoint curricula generate
safe supervision, but the deployed controller keeps only raycast and
goal-direction inputs. A proximal policy optimization (PPO) backbone
learns a compact heading-speed command~\cite{schulman2017ppo};
frozen-backbone rollouts under planner or waypoint teachers then
provide behavior-cloning data for a recurrent local
planner~\cite{bain1995framework,ross2011dagger,chen2020lbc}. The
result is a compact steering layer that preserves the learned
locomotion interface while adding memory-based decisions about when to
bypass an obstacle and when to recover the sparse goal.

LP-NavOA makes three contributions:
\begin{itemize}
\item We design a planner-free limited-perception navigation interface
that uses local range observations and goal state to steer a frozen
whole-body locomotion foundation policy through heading commands.
\item We introduce a planner-supervised training pipeline that combines
circular heading-speed PPO, waypoint/planner supervision, dynamic route
shaping, and teacher-active rollouts to distill obstacle bypassing and
post-avoidance goal recovery into the deployment policy.
\item We validate the system in MuJoCo open-wall and indoor layouts with
timing, brush/contact, and collision diagnostics, and include a Unitree
G1 deployment-feasibility check of the compact interface.
\end{itemize}

\section{Method}
\label{sec:method}

\subsection{Framework Overview: Planner-Supervised Training and Planner-Free Inference}
\label{sec:method:overview}

\begin{figure*}[!t]
\centering
\includegraphics[width=0.92\textwidth]{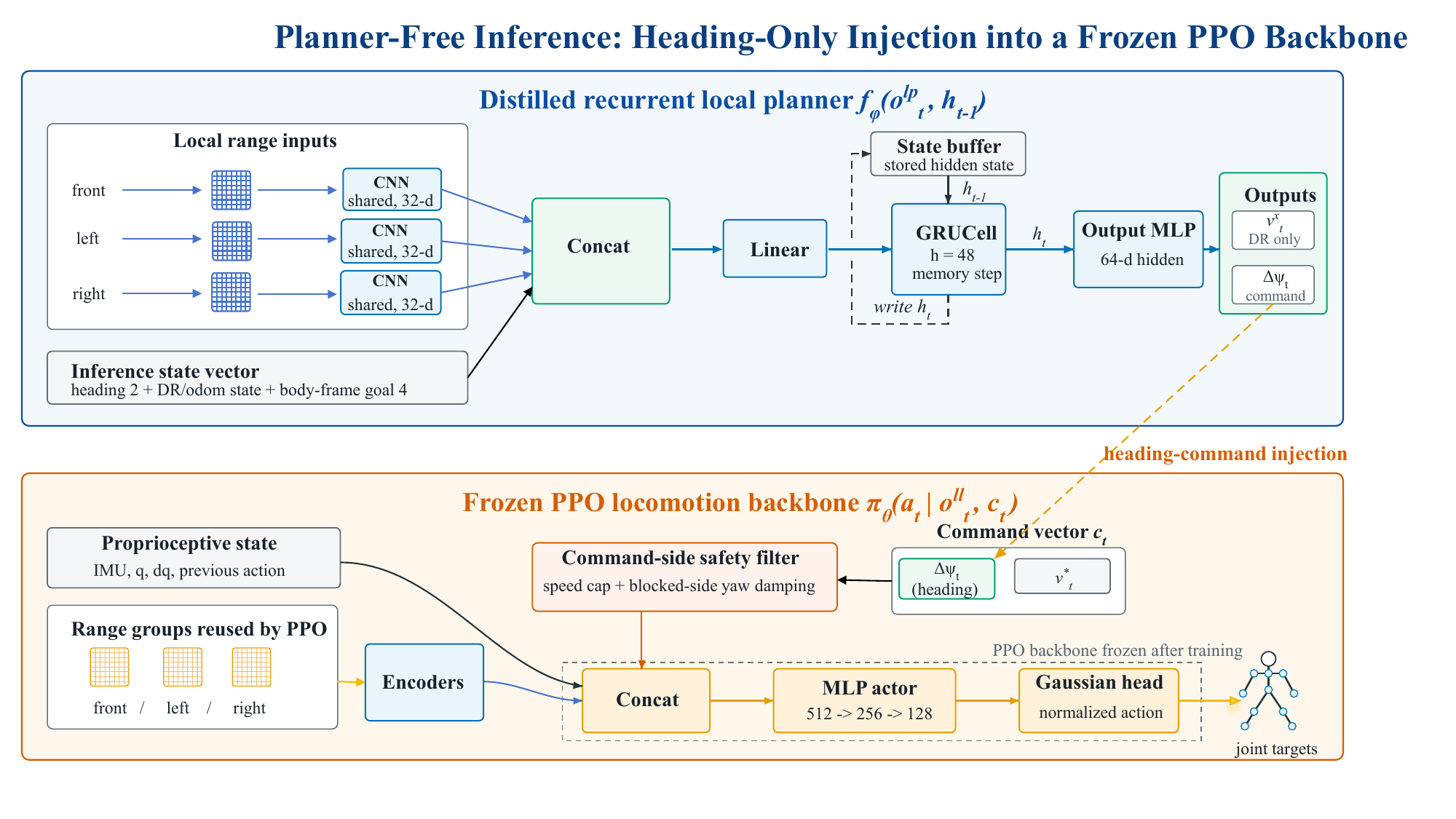}
\caption{Planner-free inference interface of LP-NavOA. The recurrent
steering module overwrites only the heading command, while the frozen PPO
backbone converts the filtered command and local observations into
full-body joint targets.}
\label{fig:arch}
\end{figure*}

We consider a Unitree G1 humanoid in MuJoCo that must reach sparse
planar goals in unknown static-obstacle layouts using only
proprioception, short-range raycasts, and relative goal direction at
deployment. Three information boundaries define the system. The locomotion
backbone
\(\pi_\theta(a_t\!\mid\!o^{\mathrm{ll}}_t,c_t)\) maps proprioception,
raycasts, and \(c_t=[\dyawt,v^\ast_t]\) to normalized joint action \(a_t\).
Training-time teachers provide intermediate waypoints for curricula and
data collection. A
recurrent local planner
\(f_\phi(o^{\mathrm{lp}}_t,h_{t-1})\) predicts a bounded steering command
from inference-time observations and recurrent memory \(h_{t-1}\), then
overwrites only the heading component. The two learning stages are
\begin{equation}
\theta^\star\!=\!\arg\max_\theta
\mathbb E_{\pi_\theta}\!\left[\textstyle\sum_t\gamma^t r_t^{\mathrm{ll}}\right],
\quad
\phi^\star\!=\!\arg\min_\phi
\mathbb E_{\mathcal D_{\pi_\theta}}\!\left[\ell(f_\phi,y_t)\right],
\end{equation}
where \(\theta\) and \(\phi\) are backbone and planner parameters,
\(\gamma\) is the discount factor, \(r_t^{\mathrm{ll}}\) is the
locomotion reward, and \(\mathcal D_{\pi_\theta}\) is collected by
rolling out the frozen backbone under teachers. Each dataset sample
contains \(o^{\mathrm{lp}}_t\), recurrent state, and teacher label
\(y_t\), equal to \(\dyawt^\ast\) for odometry and
\([\dyawt^\ast,v^x_t]\) for dead reckoning, where \(v^x_t\) is forward
body-frame velocity; \(\ell\) is the masked behavior-cloning error. The first
stage learns \emph{how} to walk under commands; the second learns
\emph{which} heading command to send without planner waypoints.
Fig.~\ref{fig:arch} summarizes this deployment
interface: planner and waypoint information is absent at runtime, the
local planner reads only deployment-available observations and recurrent
state, and its output modifies the heading component before the shared
command-side safety filter passes the command to the frozen PPO backbone.

\subsection{Perception-Action PPO Training for the Locomotion Backbone}
\label{sec:method:rl}

The first stage trains a perception-action locomotion policy rather than
a global planner. The actor maps proprioception, three local raycast
views, and the compact command to whole-body joint targets:
\begin{equation}
o^{\mathrm{ll}}_t=
\bigl[\omega^b_t,g^b_t,c_t,q_t-q_0,\dot q_t,a_{t-1},b_t,
\rho^f_t,\rho^\ell_t,\rho^r_t\bigr],
\end{equation}
where superscripts \(b\) and \(w\) denote body and world frames,
\(\omega^b_t\) is body angular velocity, \(g^b_t\) is gravity projected
into the body frame, \(q_t-q_0\) and \(\dot q_t\) are joint offsets and
velocities, \(a_{t-1}\) is the previous normalized action, and \(b_t\) is
the gait/order phase feature. The tensors
\(\rho^f_t,\rho^\ell_t,\rho^r_t\in[0,1]^{8\times8}\) are front, left,
and right normalized raycast grids; smaller entries are nearer obstacles
and 1 is the 4\,m range limit. The PD controller receives
\(q^{\mathrm{des}}_t=q_0+s_a\odot a_t\), with nominal pose \(q_0\),
per-joint scale \(s_a\), elementwise product \(\odot\), and normalized
joint action \(a_t\in[-1,1]^D\), where \(D\) is the action dimension. The critic is asymmetric and additionally
observes a yaw-aligned downward terrain scan during training.

\subsubsection{Circular goal command}
The command must teach smooth goal-directed locomotion before navigation
teachers are introduced. We therefore sample the target heading around
the current robot yaw,
\begin{equation}
\begin{aligned}
\theta^{w,\mathrm{cmd}}_t &\leftarrow
\mathrm{wrap}_{[-\pi,\pi]}\bigl(\psi_t+
U(-\theta_{\max},\theta_{\max})\bigr),\\
p^{w,\mathrm{goal}}_t &=p^w_t+
R[\cos\theta^{w,\mathrm{cmd}}_t,\sin\theta^{w,\mathrm{cmd}}_t],
\end{aligned}
\label{eq:circular-command}
\end{equation}
and the policy receives
\([\dyawt,v^\ast_t]\) with
\(\dyawt=\mathrm{wrap}(\theta^{w,\mathrm{cmd}}_t-\psi_t)\). Here
\(\psi_t\) and \(p^w_t\) are robot yaw and planar world position,
\(p^{w,\mathrm{goal}}_t\) is the sampled goal point, \(U(a,b)\) is a
uniform sample, \(R\) is the goal radius, and \(\theta_{\max}\) bounds
the angular spread around the robot heading. The wrap operation makes
\(\dyawt\) the signed shortest heading error in \([-\pi,\pi]\). This converts sparse goal tracking into
heading-and-speed tracking; obstacle curricula later reuse the same
two-dimensional
interface~\cite{bengio2009curriculum}.

\subsubsection{Rewards, safety filter, and teacher-active curricula}
Tracking targets are
\(\omega^{\ast}_t\!=\!\mathrm{clip}(k_\psi\dyawt,-\omega_{\max},\omega_{\max})\)
and
\(v^{\parallel}_t\!=\!v^x_t\cos\dyawt+v^y_t\sin\dyawt\),
with exponential rewards for yaw-rate and goal-directed velocity
tracking. Here \(k_\psi\) maps heading error to yaw-rate command,
\(\omega_{\max}\) is the yaw-rate limit, \((v^x_t,v^y_t)\) is planar base
velocity in the heading frame, \(v^{\parallel}_t\) is its component along
the commanded goal direction, and \(v^\ast_t\) is the target speed. A
command-side safety filter caps forward speed from front-ray
stopping time and damps unsafe yaw commands. It is shared by all
inference conditions, so comparisons measure steering rather than
low-level emergency logic. The reachable-goal curriculum is staged:
flat circular-goal locomotion first learns the command interface,
A-star-guided visual generalization then introduces obstacle
bypassing through intermediate waypoints, and an optional final stage
adapts the same policy to rougher terrain. Across these stages,
teachers may change the commanded heading, but the policy still sees
only \([\dyawt,v^\ast_t]\).

\subsection{Teacher-Distilled Local Planner for Goal Recovery}
\label{sec:method:lp}

After the backbone is frozen, the local planner chooses the heading
command from goal direction, local raycasts, and short-horizon
memory~\cite{kaelbling1998pomdp}. It addresses the gap left by pure
reactive avoidance: deciding when to bypass and when to return toward
the sparse goal. The student observes only inference-time signals,
\begin{equation}
\begin{aligned}
o^{\mathrm{lp}}_t=
\bigl[\eta_t,\nu_t,\rho^f_t,\rho^\ell_t,\rho^r_t,\xi_t\bigr],\\
\xi_t=\bigl[d^x_{b,t}/10,d^y_{b,t}/10,\cos\beta_t,\sin\beta_t\bigr].
\end{aligned}
\end{equation}
where \(\eta_t=[\cos\psi_t,\sin\psi_t]\) encodes yaw, \(\nu_t\) is
selected joint velocities (DR) or measured body-frame velocity
(odometry), and \(\xi_t\) stores the body-frame offset to the current key
goal, divided by 10, plus unit bearing \((\cos\beta_t,\sin\beta_t)\), with
\(\beta_t=\mathrm{atan2}(d^y_{b,t},d^x_{b,t})\). Raycast features are
fused with the state by a multilayer perceptron (MLP) and gated
recurrent unit (GRU)~\cite{cho2014gru}:
\begin{equation}
\begin{aligned}
z^s_t &= E_\rho(\rho^s_t),\quad s\in\{f,\ell,r\},\\
h_t &= \mathrm{GRUCell}\bigl(\varphi([\eta_t,\nu_t,z^f_t,z^\ell_t,z^r_t,\xi_t]),h_{t-1}\bigr),\\
\widehat{\dyawt} &= \dyaw_{\max}\tanh(W_\psi h_t).
\end{aligned}
\end{equation}
Here \(s\) indexes front/left/right raycast groups, \(E_\rho\) is the
shared CNN encoder, and \(z^s_t\) is its 32-D range feature, \(\varphi\)
is the fusion MLP, \(h_t\in\mathbb R^{48}\)
is recurrent memory, and \(W_\psi\) is the yaw-output head. The tanh bound
limits \(\widehat{\dyawt}\) to \([-\dyaw_{\max},\dyaw_{\max}]\), where
\(\dyaw_{\max}\) is the maximum steering angle. The dead-reckoning variant
also predicts forward velocity for goal integration; the odometry variant
removes that head.

\subsubsection{Route-shaping teacher control}
\begin{figure*}[!t]
\centering
\includegraphics[width=0.86\textwidth]{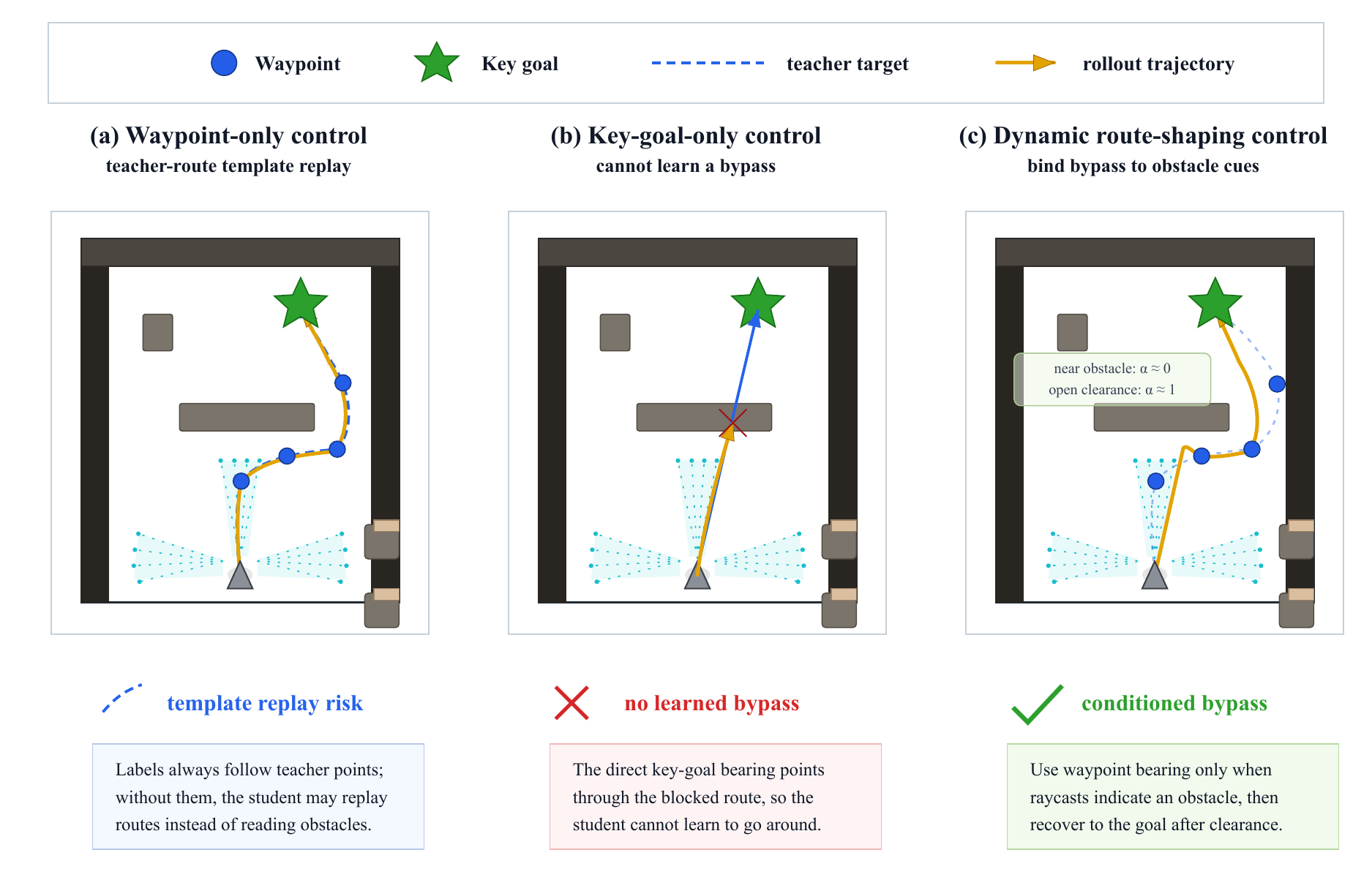}
\caption{Route-shaping teacher control in the open-wall setting.
Waypoint-only supervision can become route-template replay when teacher
points are unavailable, while key-goal-only supervision gives no bypass
target. The dynamic target ties bypassing to observed obstacle clearance
and then releases the command toward sparse-goal recovery.}
\label{fig:target-labels}
\end{figure*}

For distillation, the frozen backbone is rolled out under planner or
waypoint teachers. The active waypoint is the intermediate bypass target;
the key goal is the sparse task goal recovered after avoidance. The
teacher target blends their signed body-frame bearings:
\begin{equation}
\dyawt^\ast=\alpha_t\dyawt^{\mathrm{key}}+
(1-\alpha_t)\dyawt^{\mathrm{wp}}.
\label{eq:dynamic-label}
\end{equation}
The gate \(\alpha_t\in[0,1]\) uses the same raycasts: \(\alpha_t=0\)
selects the bypass waypoint and \(\alpha_t=1\) selects the key goal. Let
\(d_{\max}=4\,\mathrm{m}\), \(d^f_t=d_{\max}\min\rho^f_t\) be forward
clearance, and \(d^{\mathrm{gs}}_t\) the minimum left/right clearance on
the key-goal side. In the default distance-based mode,
\begin{equation}
\tilde\alpha_t=
\sigma\!\left(k_\alpha(d^f_t-d_{\mathrm{turn}})\right),
\quad
\alpha^{\mathrm{raw}}_t=
\begin{cases}
0, & d^{\mathrm{gs}}_t<d_{\mathrm{block}},\\
\tilde\alpha_t, & \mathrm{otherwise},
\end{cases}
\end{equation}
where \(\sigma\) is the logistic sigmoid, \(k_\alpha\) is transition
sharpness, \(d_{\mathrm{turn}}\) is the release clearance, and
\(d_{\mathrm{block}}\) is the goal-side blocking threshold.
The raw gate is smoothed with an asymmetric EMA
\(\alpha_t=\alpha_{t-1}+\mu_t(\alpha^{\mathrm{raw}}_t-\alpha_{t-1})\),
\(\mu_t\!=\!\mu_{\mathrm{up}}\) if
\(\alpha^{\mathrm{raw}}_t\!>\!\alpha_{t-1}\) else
\(\mu_{\mathrm{down}}\), which reduces label chatter. The gate
\(\alpha_t\) is close to zero when the waypoint should dominate near an
obstacle and moves toward one when clearances permit direct key-goal
recovery. These
quantities define both the supervised yaw target and, by default, the
heading command injected into the frozen backbone during data
collection. Thus \(\alpha_t\) shapes the rollout route rather than
merely renaming labels; the student never observes \(\alpha_t\) and is
trained with masked recurrent behavior cloning.
Fig.~\ref{fig:target-labels} explains the resulting supervision
trade-off. Waypoint-only control keeps the label tied to teacher
waypoints, so the student can replay a teacher-route template when those
points are absent instead of grounding avoidance in the observed
obstacle geometry. Key-goal-only control points through blocked space
and provides no bypass supervision. Dynamic route shaping addresses both
failure modes by using waypoint bearing only as an obstacle-conditioned
bypass cue and releasing toward the sparse goal when clearance opens.

\section{Experiments}
\label{sec:exp}

\subsection{Command-Form Ablation for Fast Backbone Training}
\label{sec:exp:goal-ablation}

Before evaluating navigation, we test whether the command interface can
train the fast locomotion primitive that the later planner relies on.
The flat-terrain PPO ablation changes only the goal-command generator:
it tests whether fixed world-frame waypoints can train the high-speed
tracking skill needed by the later navigation stack.

The \emph{circular} condition uses the robot-centered command in
\eqref{eq:circular-command}, so the target remains a local heading
and speed command as the robot moves. The \emph{fixed-sequence}
condition replaces it with four world-frame goals at nominal offsets
\((0,2.5)\), \((0,5.0)\), \((0,7.5)\), \((0,10.0)\)\,m, each scattered by
approximately \(\pm2.0\)\,m laterally and \(\pm0.5\)\,m longitudinally at
reset. Both conditions share the reward body, runner, 8192
environments, 1600 PPO iterations, and a three-stage velocity curriculum
ending at \(v_{\max}=3.0\)\,m/s, where \(v_{\max}\) is the final
commanded-speed cap.

Table~\ref{tab:goal-command-ablation} reports reward-independent
metrics over three training seeds: velocity RMSE along the active goal
direction, rollout completion, hard termination rate, and the largest
speed bin with mean velocity error below 0.5\,m/s and fall rate below
5\%.

\begin{table}[!t]
\caption{Command-form ablation for fast backbone training at
\(v_{\max}=3.0\)\,m/s. Entries are mean \(\pm\) std over three training
seeds and report reward-independent velocity tracking, rollout
completion, hard terminations, and the largest trackable speed bin.}
\label{tab:goal-command-ablation}
\begin{center}
\small
\setlength{\tabcolsep}{3pt}
\renewcommand{\arraystretch}{1.10}
\begin{tabularx}{\columnwidth}{@{}Y c c c c@{}}
\toprule
Command &
\tabhead{Vel.\ RMSE\\(m/s)} &
\tabhead{Rollout\\compl. (\%)} &
\tabhead{Fall\\rate (\%)} &
\tabhead{Trackable\\speed (m/s)} \\
\midrule
Circular moving goal & \best{0.52$\pm$0.02} & \best{99.51$\pm$0.15} & 0.49$\pm$0.15 & \best{3.00$\pm$0.00} \\
Fixed scattered sequence & 1.36$\pm$0.03 & 40.07$\pm$2.95 & \best{0.09$\pm$0.04} & 1.44$\pm$0.10 \\
\bottomrule
\end{tabularx}
\end{center}
\end{table}

The circular command remains trackable at 3.0\,m/s with 0.52\,m/s RMSE
and 99.51\% rollout completion. The fixed sequence rarely falls, but
its trackable speed is only 1.44\,m/s and only 40.07\% of rollouts
complete, showing that the circular command is a training interface for
the fast backbone rather than a notational convenience.

\subsection{Simulation Evaluation of Obstacle Avoidance and Goal-Directed Navigation}
\label{sec:exp:setup}
\label{sec:exp:results}

The formal simulation benchmark evaluates the frozen backbone with
planner-assisted teachers and planner-free local-planner inference in
the same MuJoCo G1 simulator. The official navigation scope is two
obstacle families:
\textbf{open straight-wall}, where the robot approaches a wall at
0.5--1.2\,m/s and must bypass it, and \textbf{indoor obstacle layout},
with room boundaries, a central wall, and box obstacles.

\subsubsection{Policy conditions and metrics}
\textbf{R1 (Backbone-only)} sends the heading command directly toward
the active key goal and acts as a sanity-check lower bound in
wall-obstructed scenes. \textbf{\Ours\ (DR)} uses the frozen backbone
and learned planner in planner-free (NoPP) inference with
dead-reckoning (DR) state. \textbf{\Ours\ (Odom)} uses the same steering
interface but computes the relative goal from odometry, removes the
velocity head, and excludes joint velocities. \textbf{T1 (Teacher, PP)}
keeps the teacher active under planner-present (PP) inference and is
used only as a calibration reference, not as a
method row in Table~\ref{tab:main}. All conditions share raycast
sensors, episode limits, and the same command-side safety filter.

Timeout-based final-goal completion is intentionally not used as the
success criterion: the simulator leaves enough time for a failed first
approach to circle back to the target, a behavior with little practical
value in crowded indoor spaces. We therefore evaluate whether the robot
reaches the goal on its first useful approach and without scraping
through the safety envelope. The objective reference time is
\(T_0=L_{\mathrm{obj}}/v_{\mathrm{cmd}}\), with nominal obstacle-aware
route length \(L_{\mathrm{obj}}\) and commanded speed \(v_{\mathrm{cmd}}\).
The displayed \emph{T-rel.\ on-time} metric uses the same task/seed T1
stream as calibration: successful teacher-present episodes define the
98\% threshold of
\(d_{\mathrm{eff}}=T_{\mathrm{ep}}v_{\mathrm{cmd}}\), where
\(T_{\mathrm{ep}}\) is episode time and \(d_{\mathrm{eff}}\) is the
corresponding effective distance. Thus T-rel.\ on-time is the percentage
of episodes reaching the goal within the T1-calibrated effective-distance
budget. Expressed as time, this T-rel.\ window is
about 25--26\,s for the open-wall task and 23--31\,s for the indoor task,
roughly 5--12\,s longer than the nominal best-path time depending on the
seed and layout. Table~\ref{tab:main} reports T-rel.\ on-time together
with the observed overhead \(\Delta t=T_{\mathrm{ep}}-T_0\), so the
reader can see how many seconds each method spends beyond
\(T_0\). \emph{Any brush}
combines sustained safety-envelope braking and body/foot contact brush;
\emph{contact brush} counts body or foot obstacle contact below hard
collision termination, and \emph{hard collision} is a termination-level
collision.

Table~\ref{tab:main} quantifies how much of the teacher behavior is
retained by the planner-free recurrent local planner while the frozen
whole-body controller, sensors, episode limits, and safety filter remain
fixed.

\begin{table*}[!t]
\caption{Formal v6 navigation quality on open-wall and indoor layouts.
Percentage columns are mean \(\pm\) std over seeds \(0,1,2\); the timing
column reports T-rel.\ on-time arrival with the mean overhead
\(\Delta t=T_{\mathrm{ep}}-T_0\). Boldface marks the best displayed mean
within each task block.}
\label{tab:main}
\centering
\footnotesize
\setlength{\tabcolsep}{3pt}
\renewcommand{\arraystretch}{1.00}
\begin{tabular*}{0.96\textwidth}{@{\extracolsep{\fill}}l l c c c c@{}}
\toprule
Task & Method &
\tabhead{T-rel.\\on-time / \(\Delta t\)} &
\tabhead{Any brush} &
\tabhead{Contact\\brush} &
\tabhead{Hard coll.} \\
\midrule
\multirow{3}{*}{Open wall}
  & R1 Backbone-only   & 38.3$\pm$5.2 / $+14.1$s & 70.9$\pm$7.7 & 69.7$\pm$7.4 & 0.3$\pm$0.1 \\
  & \Ours\ (DR) NoPP   & 85.1$\pm$1.3 / $+7.5$s  & 24.0$\pm$1.7 & \best{9.5$\pm$2.0} & 0.2$\pm$0.2 \\
  & \Ours\ (Odom) NoPP & \best{91.9$\pm$2.1 / $+6.1$s} & \best{23.7$\pm$1.4} & 10.1$\pm$0.8 & \best{0.1$\pm$0.1} \\
\midrule
\multirow{3}{*}{Indoor}
  & R1 Backbone-only   & 40.3$\pm$15.3 / $+14.4$s & 77.7$\pm$7.3 & 76.5$\pm$7.4 & 0.5$\pm$0.4 \\
  & \Ours\ (DR) NoPP   & 88.0$\pm$4.3 / $+7.4$s  & 21.2$\pm$2.7 & 12.4$\pm$1.9 & 0.1$\pm$0.1 \\
  & \Ours\ (Odom) NoPP & \best{96.6$\pm$2.6 / $+5.5$s} & \best{9.7$\pm$2.1} & \best{4.9$\pm$1.6} & \best{0.0$\pm$0.0} \\
\bottomrule
\end{tabular*}
\end{table*}

\begin{figure*}[!tbp]
\centering
\includegraphics[width=0.92\textwidth]{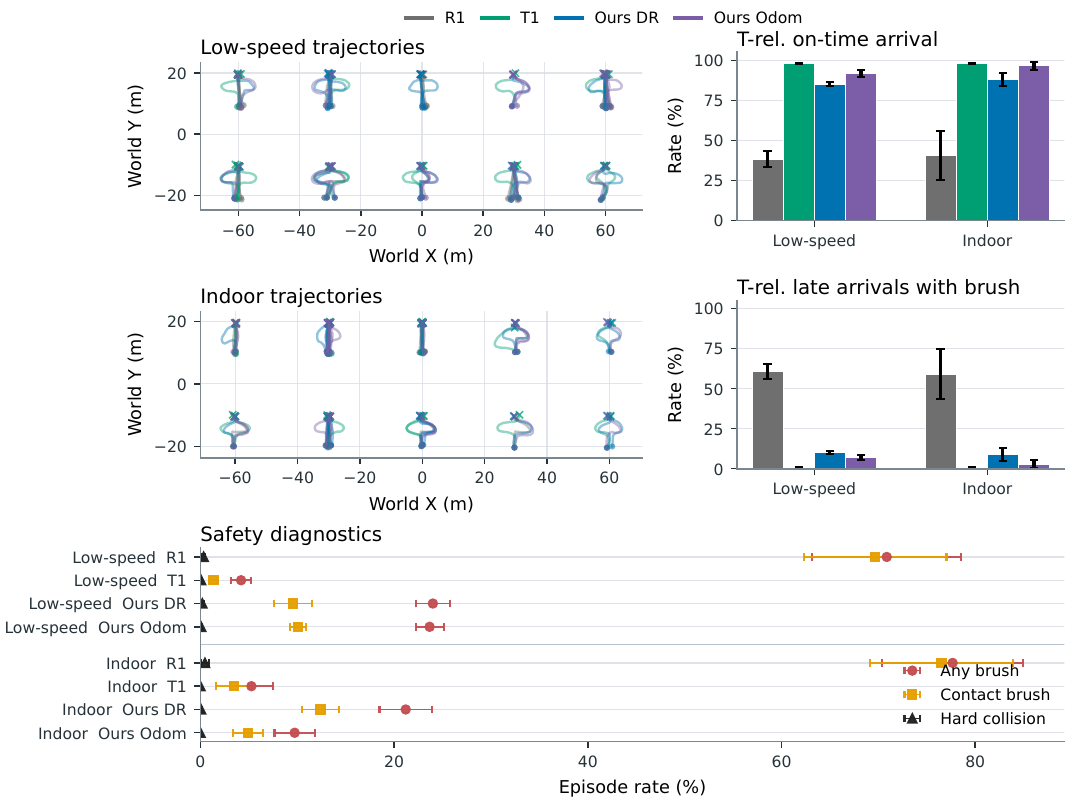}
\caption{Representative planner-free trajectories and navigation-quality
diagnostics. The figure links post-obstacle goal recovery in sample
rollouts with the aggregate timing and brush/contact trends reported in
Table~\ref{tab:main}.}
\label{fig:diagnostics}
\end{figure*}

The compact timing column exposes the main pattern. R1 often reaches
the final goal eventually, but its mean overhead is about \(+14\)\,s
beyond the nominal best-path time and only about 38--40\% of its runs
arrive within the T1-calibrated budget. This happens because the shared
safety envelope and jointly trained perception-action locomotion model
already give the backbone a weak ability to avoid and edge around
obstacles, but the behavior is indirect and frequently contact-heavy.
The learned planners reduce the overhead to about \(+5.5\)--\(+7.5\)\,s,
raising T-rel.\ on-time arrival to 85--97\%. Brush diagnostics follow
the same trend, while hard collisions remain rare; the benefit is
therefore direct, timely arrival rather than simply avoiding
catastrophic failure. Fig.~\ref{fig:diagnostics} supports this
interpretation by pairing representative planner-free trajectories with
navigation-quality diagnostics: the traces recover toward the goal after
bypassing the obstacle, and the aggregate panel shows the corresponding
gain in on-time arrival and reduction in brush/contact-heavy progress.

\subsection{Ablations on Route-Shaping Control, Odometry, and Safety Filtering}
\label{sec:exp:ablation}

\begin{figure}[!tbp]
\centering
\includegraphics[width=\columnwidth]{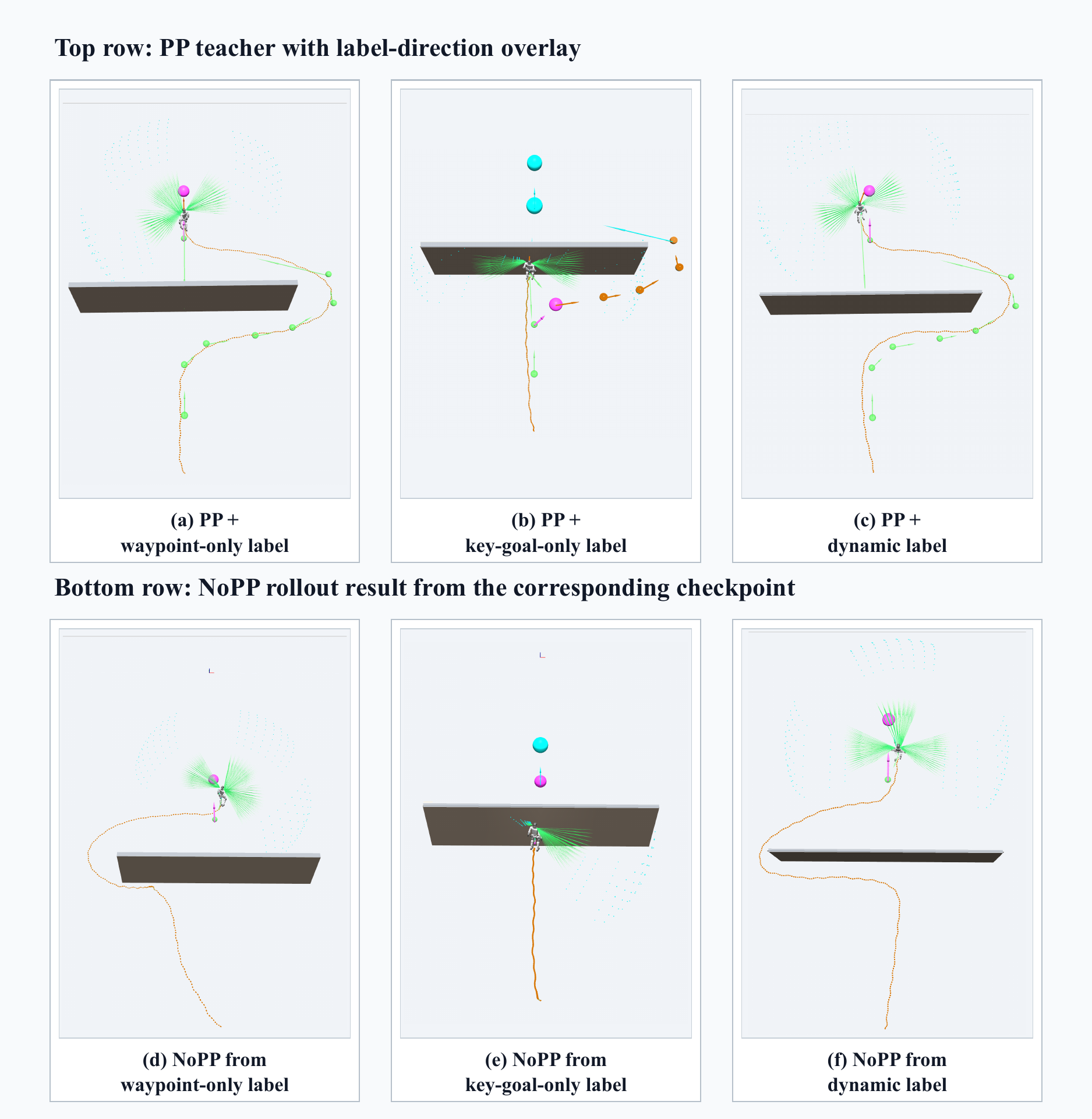}
\caption{Open-wall route-control ablation snapshots. Each panel compares
the planner/waypoint teacher rollout (PP) with the corresponding
planner-free local-planner rollout (NoPP), highlighting whether the
learned policy reproduces bypassing and goal recovery.}
\label{fig:low-speed-label-ablation}
\end{figure}

The navigation ablation uses the open straight-wall task, where the
distinction between obstacle avoidance and returning to the key goal
is easiest to attribute. For the DR variant, target-shaping ablations
compare waypoint-only and key-goal-only route targets against the
dynamic route-shaping control in \eqref{eq:dynamic-label}.
Table~\ref{tab:ablation} then quantifies the same variants for both
dead-reckoning and odometry state, including delayed arrival, brush
events, and the teacher-forced-to-planner-free transfer gap. The
``no teacher forcing'' row removes the default path in which this target
is also injected into the frozen backbone during data collection. Both
modes include a safety-filter stress test that disables the velocity
envelope on the same checkpoint. Fig.~\ref{fig:low-speed-label-ablation}
provides the qualitative counterpart to these rows: key-goal-only
control fails to learn a bypass around the occluded direct route,
waypoint-only control tends to carry the teacher-route template into
NoPP rollout, and dynamic route shaping supplies an obstacle-conditioned
bypass-and-recover pattern.

\begin{table*}[!t]
\caption{Open-wall route-control ablation over three seeds. T-rel.\
on-time uses the first-approach budget defined for Table~\ref{tab:main};
T-rel.\ late is the complementary delayed-arrival rate among completed
episodes, and PP--NoPP gap is the teacher-forced reference minus
planner-free rollout completion. Boldface marks the best mean within
each mode block.}
\label{tab:ablation}
\centering
\footnotesize
\setlength{\tabcolsep}{2pt}
\renewcommand{\arraystretch}{1.00}
\begin{tabular*}{0.96\textwidth}{@{\extracolsep{\fill}}l l c c c c c@{}}
\toprule
Mode & Variant &
\tabhead{T-rel.\\on-time} &
\tabhead{T-rel.\\late} &
\tabhead{Any brush} &
\tabhead{Hard coll.} &
\tabhead{PP--NoPP gap} \\
\midrule
DR & Full                 & 85.3 & 14.5 & 23.1 & \best{0.1} & \best{0.0} \\
DR & Waypoint-only control & 72.6 & 22.3 & 33.0 & 0.3 & 4.9 \\
DR & Key-goal-only control & 33.9 & \best{0.1} & 66.5 & 0.5 & 65.9 \\
DR & No teacher forcing   & 67.9 & 27.9 & 53.8 & \best{0.1} & 4.1 \\
DR & No safety filter     & \best{87.0} & 12.5 & \best{15.6} & 0.5 & 0.3 \\
\midrule
Odom & Full                 & 92.5 & 7.5 & 24.2 & \best{0.0} & \best{-0.1} \\
Odom & Waypoint-only control & 89.8 & 10.2 & 19.1 & \best{0.0} & \best{-0.1} \\
Odom & Key-goal-only control & 32.4 & \best{0.1} & 67.7 & \best{0.0} & 67.4 \\
Odom & No safety filter     & \best{95.7} & 4.2 & \best{16.8} & 0.1 & \best{-0.1} \\
\bottomrule
\end{tabular*}
\end{table*}

The ablation isolates three effects. Key-goal-only control fails in
both modes: only about one third of runs arrive on time, and any-brush
rates exceed 65\%. Waypoint-only control avoids the direct-target
failure mode, but its labels are anchored to teacher waypoints; once
those waypoints are removed at NoPP inference, the student can reproduce
a memorized detour template rather than condition avoidance on the
currently observed obstacle. Dynamic route shaping is more efficient
because it makes the waypoint bearing active only when local clearance
indicates that bypassing is needed and otherwise returns supervision to
the key goal. For DR, removing teacher-label-to-policy injection raises
any brush from 23.1\% to 53.8\% and increases the PP--NoPP gap from
0.0\% to 4.1\%, confirming that route shaping must affect the rollout
state distribution. The no-safety-filter rows give the best simulated
timing and brush rates, which indicates that the front local planner can
already perform strong obstacle avoidance in MuJoCo. The safety filter
is therefore retained mainly for real-robot deployment as a final
braking layer: it may make simulated motion more conservative, but it
provides the last line of defense against an unmodelled obstacle or
sensing error.

\subsection{Real-Robot Deployment Analysis}
\label{sec:exp:hardware}

\begin{figure}[!tbp]
\centering
\subfloat[Real-robot avoidance montage.]{
\includegraphics[width=0.985\columnwidth]{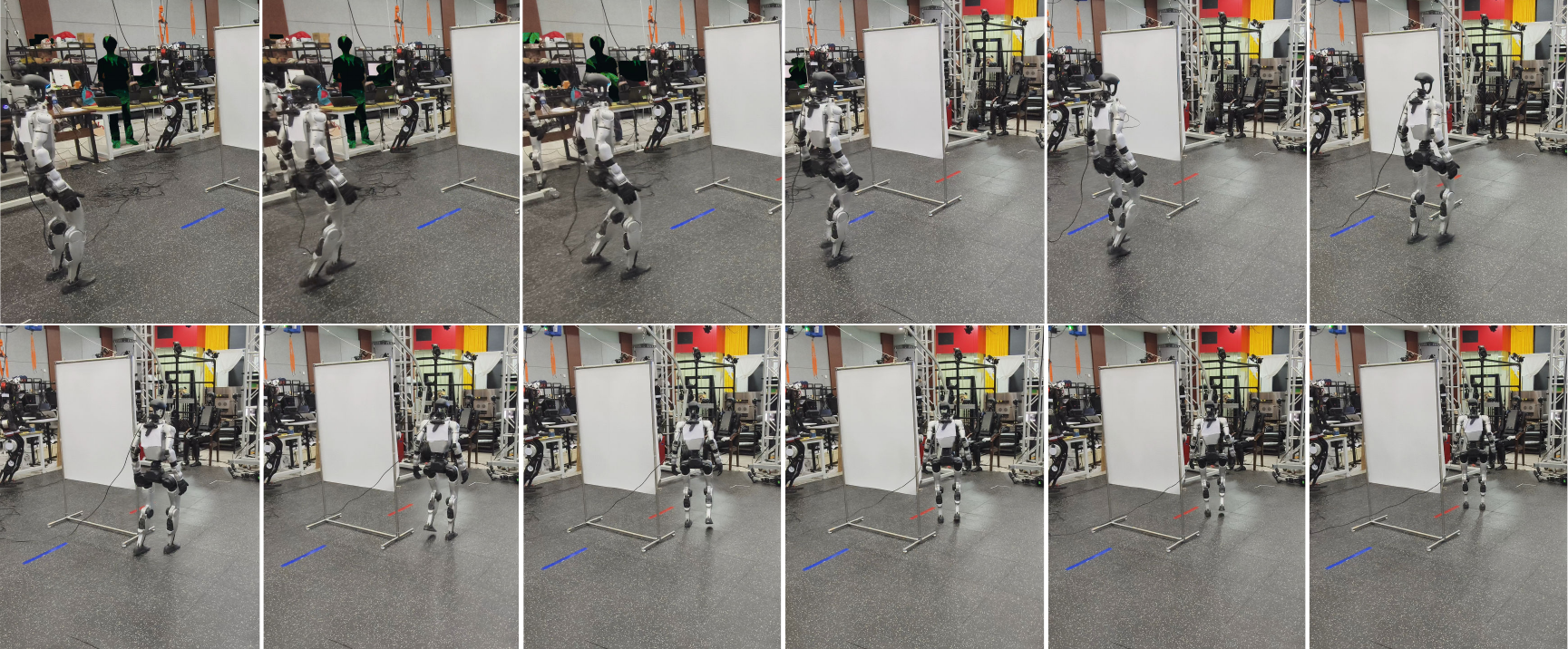}}\\[0.3em]
\begin{minipage}[t]{0.60\columnwidth}
\centering
\subfloat[Logged deployment trajectory.]{
\includegraphics[width=0.96\linewidth]{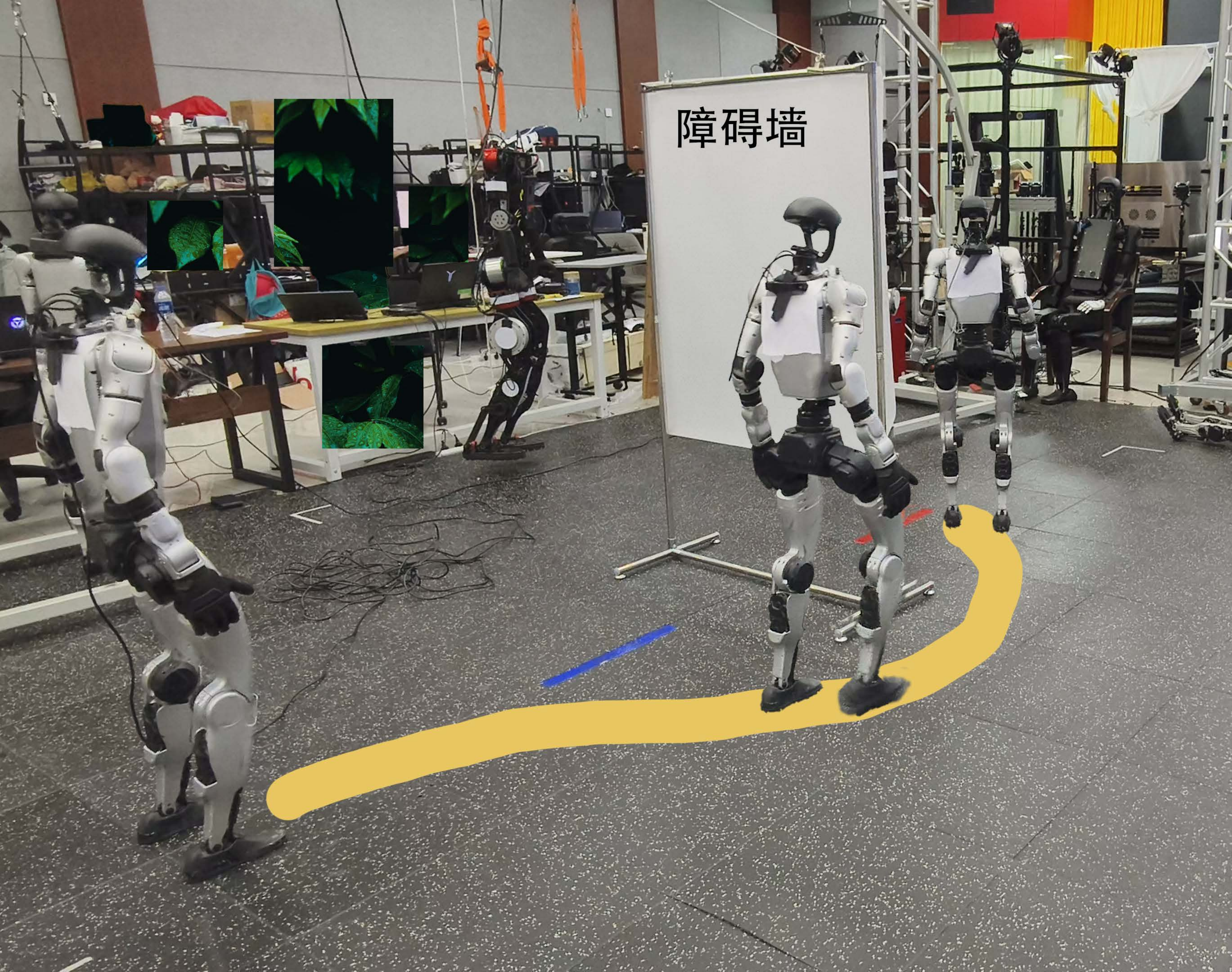}}
\end{minipage}\hfill
\begin{minipage}[t]{0.28\columnwidth}
\centering
\subfloat[Sensor.]{
\includegraphics[width=0.56\linewidth]{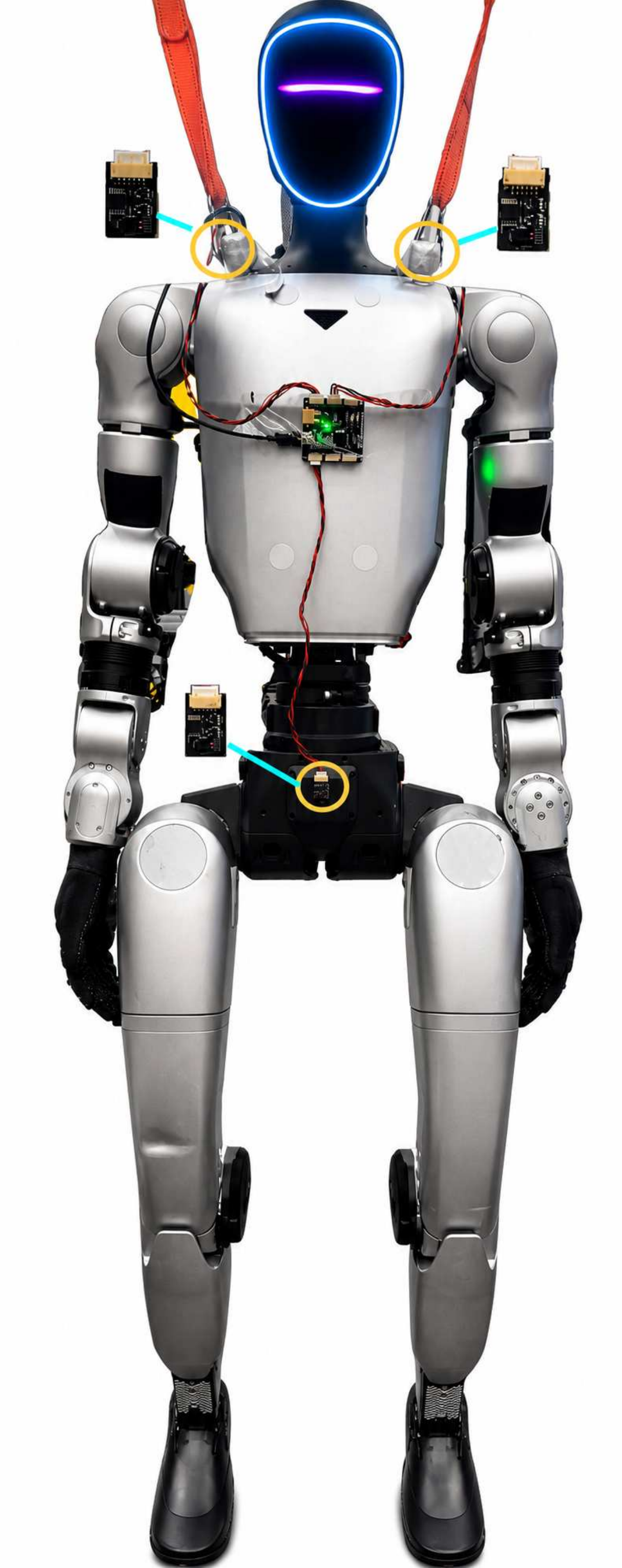}}
\end{minipage}
\caption{Real-robot feasibility validation. After receiving a
staged body-frame target and commanded speed, the planner-free steering
policy approaches the obstacle, selects a bypass direction, and returns
toward the target without continuous joystick steering.}
\label{fig:hardware-validation}
\end{figure}

We use the real-robot run as a deployment-feasibility check for the
compact inference interface. The deployed stack pairs the
raycast-conditioned locomotion backbone with the odometry-assisted front
local planner; on hardware, odometry supports target recovery while
local range observations remain in the locomotion loop. The check covers
representative wall-occlusion layouts: frontal single-wall
barriers of different lengths, an obliquely placed single-wall barrier,
and a front--rear two-wall barrier configuration. In all cases, the
commanded target is placed in front of the robot, so the policy must
first select a feasible bypass direction and then recover toward the
forward target after the occlusion is cleared. The recorded trials span
commanded speeds from 0.5 to 1.0\,m/s, and the available front-range
trace records a closest observed clearance of 0.281\,m. These deployment
logs verify that the compact inference interface can issue executable
bypass-and-recover steering commands on the Unitree G1 without
continuous joystick guidance. Fig.~\ref{fig:hardware-validation} shows
one representative montage, trajectory, and sensor record; detailed
videos are included with the open-source artifacts in the Appendix.

\section{Conclusion}
\label{sec:conclusion}

We presented LP-NavOA, a limited-perception framework that integrates
local navigation and obstacle avoidance for humanoid robots by placing a
learned recurrent steering module above a frozen raycast-conditioned
whole-body locomotion backbone. The central design is to use planners and waypoint teachers
only during training, while deployment retains a compact interface:
proprioception, local range sensing, body-frame goal direction, a
command-side safety filter, and a heading command sent to the backbone.
This separation lets the system inherit stable learned locomotion while
adding memory-based decisions about when to bypass an obstacle and when
to recover the sparse goal.

The experiments show that this integrated steering layer changes the
quality of navigation rather than merely the final timeout-based
completion rate. On open-wall and indoor layouts, the distilled planner
substantially improves T-rel.\ on-time arrival and reduces
brush/contact-heavy progress compared with the backbone-only controller,
while keeping hard collisions rare. The ablations further indicate that
the circular heading-speed command is important for training the
3.0\,m/s backbone, and that dynamic route shaping plus teacher-active
data collection are necessary for the planner-free student to reproduce
obstacle-conditioned bypass and goal-recovery behavior without relying
on explicit teacher waypoints at inference. The
no-safety-filter results suggest that the learned local planner already
handles most simulated obstacle avoidance; the command-side safety
filter is retained primarily as a conservative deployment safeguard.

The Unitree G1 deployment check confirms executable planner-free
steering on hardware; repeated trials, moving obstacles, and richer
indoor layouts remain natural extensions.

\FloatBarrier

\appendix[Implementation Details]
\label{app:details}

Table~\ref{tab:implementation-details} summarizes the reproducibility
configuration. Source code, configuration files, and supplementary
real-robot videos are available at
\url{https://shenqiqishi.github.io/LP-NavOA/}.

\begin{table}[!tbp]
\caption{Reproducibility configuration for the simulated training and
evaluation setup.}
\label{tab:implementation-details}
\begin{center}
\scriptsize
\setlength{\tabcolsep}{3pt}
\renewcommand{\arraystretch}{0.96}
\begin{tabularx}{\columnwidth}{@{}l Y@{}}
\toprule
Item & Setting \\
\midrule
Open-source base & \texttt{mjlab} manager-based RL environment,
task registry, and PPO training/evaluation scripts. \\
Robot / simulator & Unitree G1 humanoid in MuJoCo. \\
Range sensing & Three VL53L8CX-style front/left/right raycast groups,
each \(8\times8\), 45\(^\circ\) FOV, 4\,m range, base-aligned, 12\,Hz. \\
Backbone network & Actor/critic MLP \((512,256,128)\) with ELU
activations and observation normalization; each raycast group uses a
small convolutional encoder. \\
PPO training & Clip 0.2, entropy 0.01, GAE
\(\gamma=0.99,\lambda=0.95\), adaptive LR \(10^{-3}\), 5 epochs, 4
mini-batches, 24 steps per environment. \\
Goal-command ablation & 8192 parallel environments, 1600 iterations per
seed, three train seeds, velocity curriculum ending at 3.0\,m/s. \\
Navigation evaluation & Seeds \(0,1,2\), 500 episodes per seed, 256
parallel environments; formal scope limited to open straight-wall and
indoor layouts. \\
Training hardware & Intel Core i9-14900K CPU and NVIDIA GeForce RTX
4090D GPU. \\
\bottomrule
\end{tabularx}
\end{center}
\end{table}

\FloatBarrier

Table~\ref{tab:reward-details} lists the principal locomotion reward
terms; navigation comparisons keep the backbone and safety filter fixed.

\FloatBarrier

\begin{table}[!tbp]
\caption{Principal locomotion reward terms used to train the G1
backbone.}
\label{tab:reward-details}
\begin{center}
\scriptsize
\setlength{\tabcolsep}{3pt}
\renewcommand{\arraystretch}{0.96}
\begin{tabularx}{\columnwidth}{@{}l c @{\hspace{2em}} Y@{}}
\toprule
Term & Weight & Role \\
\midrule
\texttt{track\_ang\_vel\_cmd} & 1.3--1.8 & yaw-rate command tracking \\
\texttt{track\_goal\_velocity} & 2.0 & goal-directed speed tracking \\
\texttt{flat\_orientation\_l2} & -5.0 & upright torso regulation \\
\texttt{is\_terminated} & -200 & fall or illegal termination penalty \\
\texttt{joint\_pos\_limits} & -10 & joint-limit avoidance \\
\texttt{action\_rate\_l2} & -0.05 & smooth joint targets \\
\texttt{foot\_slip} & -0.25 & foot-slip suppression \\
\texttt{body\_collision} & -1.0 & obstacle/body contact penalty \\
\texttt{feet\_collision} & -0.5 & horizontal foot-impact penalty \\
\texttt{in\_place\_turn\_bonus} & +1.0 & large heading-error turning \\
\bottomrule
\end{tabularx}
\end{center}
\end{table}

\FloatBarrier


\bibliographystyle{IEEEtran}
\bibliography{ref}

\end{document}